\documentclass[letterpaper]{article} 
\usepackage[submission]{aaai2027}  
\usepackage[hyphens]{url}  
\usepackage{graphicx} 
\usepackage{natbib}  
\usepackage{caption} 
\usepackage{algorithm}
\usepackage{algorithmic}
\usepackage{amsfonts} 
\usepackage{amsmath}
\usepackage{multirow} 
\usepackage{pifont}

\usepackage[table]{xcolor}
\usepackage{amssymb}

\usepackage{newfloat}
\usepackage{listings}
\DeclareCaptionStyle{ruled}{labelfont=normalfont,labelsep=colon,strut=off} 
\floatstyle{ruled}
\newfloat{listing}{tb}{lst}{}
\floatname{listing}{Listing}

\usepackage{booktabs}

\title{DSCD-Nav: Dual-Stance Cooperative Debate for Zero-Shot Object Navigation}
\author{
    Weitao An\textsuperscript{1},
    Qi Liu\textsuperscript{1},
    Chenghao Xu\textsuperscript{2},
    Jiayi Chai\textsuperscript{1},
    Xu Yang\textsuperscript{1},
    Kun Wei\textsuperscript{1},
    Cheng Deng\textsuperscript{1}
}
\affiliations{
    \textsuperscript{1}School of Electronic Engineering, Xidian University, Xi'an 710071, China\\
    \textsuperscript{2}School of Information Science and Engineering, Hohai University, Nanjing 210098, China
}

\begin{document}

\maketitle

\begin{abstract}

Adaptive navigation in unfamiliar indoor environments is crucial for household service robots. Despite advances in zero-shot perception and reasoning from vision–language models, existing navigation systems still rely on single-pass scoring at the decision layer, leading to overconfident long-horizon errors and redundant exploration. To tackle these problems, we propose Dual-Stance Cooperative Debate Navigation (\textbf{DSCD-Nav}), a decision mechanism that replaces one-shot scoring with stance-based cross-checking and evidence-aware arbitration to improve action reliability under partial observability. Specifically, given the same observation and candidate action set, we explicitly construct two stances by conditioning the evaluation on diverse and complementary objectives: a Task–Scene Understanding (TSU) stance that prioritizes goal progress from scene-layout cues, and a Safety–Information Balancing (SIB) stance that emphasizes risk and information value. The stances conduct a cooperative debate and make policy by cross-checking their top candidates with cue-grounded arguments. Then, a Navigation Consensus Arbitration (NCA) agent is employed to consolidate both sides’ reasons and evidence, optionally triggering lightweight micro-probing to verify uncertain choices, preserving NCA's primary intent while disambiguating. Experiments on HM3Dv1, HM3Dv2, and MP3D demonstrate consistent improvements in success and path efficiency. Real-world experiments further demonstrate its effectiveness in robot navigation.

\end{abstract}


\section{Introduction}

Indoor service robots require robust navigation to reach target
objects in complex environments. In zero-shot object navigation,
an embodied agent must reach an unseen target using only egocentric
observations, without prior maps or task-specific training.
Given only partial views and an object-level goal, the agent must
interpret observations, explore informative viewpoints, and make
long-horizon decisions under incomplete evidence.
\begin{figure}[t]
  \begin{center}
    \centerline{\includegraphics[width=\columnwidth]{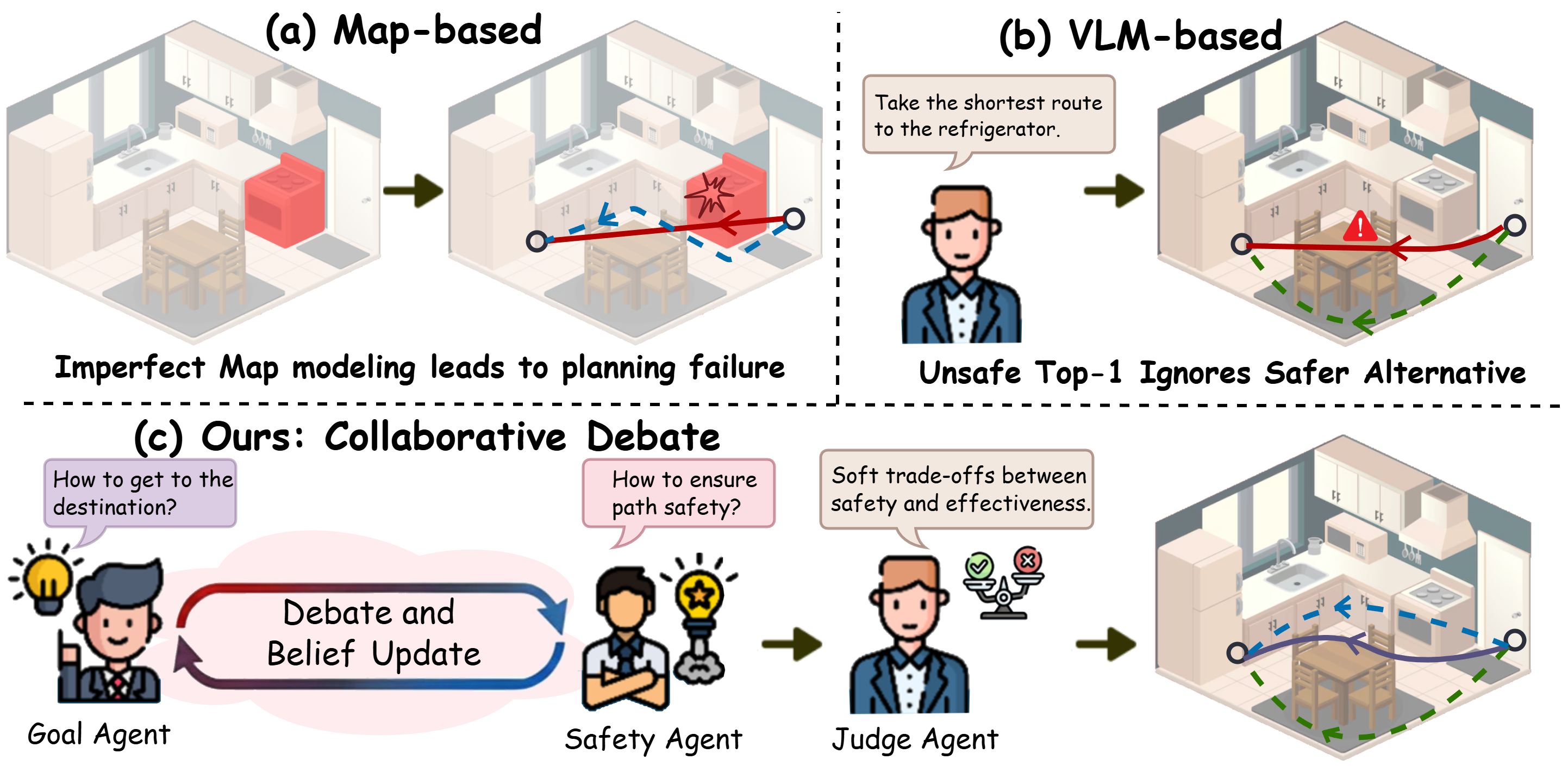}}
    \caption{
      (a) Map-based navigation is brittle to map errors. (b) VLM-based navigation is overly confident, making greedy Top-1 choices. (c) Our DSCD-Nav uses collaborative debate and arbitration for more reliable decisions.
    }
    \label{fig:intro_overview}
  \end{center}
\end{figure}

Existing indoor navigation systems typically evolve from explicit mapping-and-planning pipelines to foundation-model-driven decision makers. Early approaches build structured spatial representations, including globally consistent topological graphs, semantic maps, or online occupancy grids, to support downstream planning and control ~\cite{11128393,Jiang_2025,zhang-etal-2025-mapnav,NEURIPS2024_098491b3}. When geometry and semantics are reliable, these map-centric pipelines can select waypoints efficiently and generate stable executable paths. More recently, to reduce reliance on precise maps and large-scale task training, LLMs and VLMs have been incorporated into navigation systems to enhance scene understanding and goal grounding ~\cite{goetting2024endtoendnavigationvisionlanguage,long2024instructnav}. For example, an LLM can translate a goal or instruction into semantic sub-goals, while a VLM can directly score candidate directions or views from the current observation, enabling strong zero-shot behaviors ~\cite{nasiriany2024pivot}.

Despite their effectiveness, both paradigms remain brittle in complex indoor scenes, as action selection often commits to a single stream, slow to challenge or revise under partial evidence. In map-centric pipelines, occlusions, lighting variation, or clutter may introduce systematic errors into the representation, yet the planner may still trust it and execute numerous steps before reacting ~\cite{semantic_mapping}. As illustrated in Fig.~\ref{fig:intro_overview}(a), such over-commitment can trigger prolonged detours (\textit{e.g.}, mistaking a corridor for a dead end) or inefficient exploration. VLM-based agents reduce mapping, but many still make one-shot decisions by scoring candidates once and executing the top-ranked action ~\cite{gu2025doraemondecentralizedontologyawarereliable,11246684} as shown in Fig.~\ref{fig:intro_overview}(b). In both cases, goal progress, uncertainty, and reliability are often compressed into a single scalar score ~\cite{du2025vlnavrealtimevisionlanguagenavigation}, hindering conflicting evidence discovery and early correction of premature commitments, even with VLM backbones ~\cite{Zhou_Hong_Wu_2024}. These limitations suggest that reliable navigation requires explicit evidence cross-checking before execution to mitigate overconfidence and weak self-correction in single-stream navigation pipelines. 
Multiple reasoning perspectives offer a natural solution, yet homogeneous agents may still converge to the same mistaken decision when their reasoning patterns are overly similar.

Motivated by analyses showing that homogeneous debate may reinforce shared misconceptions and that symmetric interaction does not necessarily produce corrective updates \cite{NEURIPS2024_32e07a11,choi2025debatevoteyieldsbetter}, we propose Dual-Stance Cooperative Debate Navigation (DSCD-Nav), a decision mechanism that replaces conventional one-shot scoring with stance-based cross-checking and evidence-aware arbitration. Instead of increasing the number of homogeneous agents, DSCD-Nav introduces two complementary yet heterogeneous epistemic stances over a shared candidate action set: the Task--Scene Understanding (TSU) stance maintains a goal-directed action hypothesis from scene and target consistency, while the Safety--Information Balancing (SIB) stance performs evidence-critical verification by examining risk, traversability, visibility, and information value. Their structured disagreement enables corrective updates rather than simple consensus formation. A Navigation Consensus Arbitration (NCA) agent then consolidates the debate trace and selects the final action, optionally triggering lightweight micro-probing when residual uncertainty requires additional observations. By explicitly exposing candidate conflicts and preserving reliable decisions, DSCD-Nav improves action reliability under partial observability.
Overall, the main contributions of this work are as follows:

\begin{itemize}
    \item
    We introduce DSCD-Nav, a training-free dual-stance decision
    framework for zero-shot object navigation that replaces
    one-shot candidate ranking with structured evidence-conflict
    resolution under partial observability.
    DSCD-Nav is designed as a modular decision verification layer
    that can interface with existing candidate generation pipelines.
    
    \item
    We design a heterogeneous and asymmetric cooperative debate protocol with complementary and non-exchangeable reasoning roles over a
    shared executable candidate set.
    Candidate-grounded counter-evidence, revision-trace-aware NCA
    arbitration, and bounded micro-probing jointly expose conflicting
    evidence, support corrective preference revisions, and acquire
    new observations under unresolved local ambiguity.
        
    \item Extensive experiments on HM3Dv1, HM3Dv2, and MP3D demonstrate consistent improvements over existing zero-shot navigation approaches, with real-world robot evaluations further validating the effectiveness and deployability of DSCD-Nav.

\end{itemize}

\section{Related Work}
\subsection{Map-Centric Navigation and Frontier Exploration}
Classical embodied navigation methods rely on explicit geometric and semantic representations, including occupancy maps \citep{semantic_mapping}, topological graphs \citep{Yang_2025_CVPR}, and open-vocabulary semantic maps \citep{kuang-etal-2024-openfmnav,11128393,Jiang_2025,zhang-etal-2025-mapnav}, to support interpretable exploration and goal-directed search in unseen environments \citep{Al-Halah_2022_CVPR,Xu_2024_CVPR,Ziyue_WANG20252025PCP0002}. Online scene graphs \citep{NEURIPS2024_098491b3} and frontier-based reasoning \citep{10610712,ramakrishnan2025does} further improve search efficiency, but these pipelines remain sensitive to localization errors, semantic projection noise \citep{semantic_mapping,xu2025embodiedsam}, and handcrafted scoring. Consequently, representation bias propagates to downstream decisions without action-level self-correction.

\subsection{VLM and World-Model-Enhanced Navigation}
Recent work brings VLM understanding \citep{goetting2024endtoendnavigationvisionlanguage,Yin_2025_CVPR} into candidate-action selection via iterative prompting \citep{nasiriany2024pivot}, and augments decisions with world models \citep{11246684} or future-view prediction \citep{zhang2024navidvideobasedvlmplans} for planning \citep{Wang_2023_ICCV}. These designs improve evidence–goal alignment \citep{rajabi2025travel} and robustness in long-horizon settings via spatial reasoning \citep{zhong2025topvnavunlockingtopviewspatial,du2025vlnavrealtimevisionlanguagenavigation} and memory \citep{Cao_2025_ICCV,zeng2025janusvlndecouplingsemanticsspatiality,11178223}. Still, action choice often collapses to a single best candidate \citep{gu2025doraemondecentralizedontologyawarereliable}, leaving conflicts and trade-offs implicit and allowing early bias to accumulate.

\begin{figure*}[ht]
  \begin{center}
    \centerline{\includegraphics[width=\textwidth]{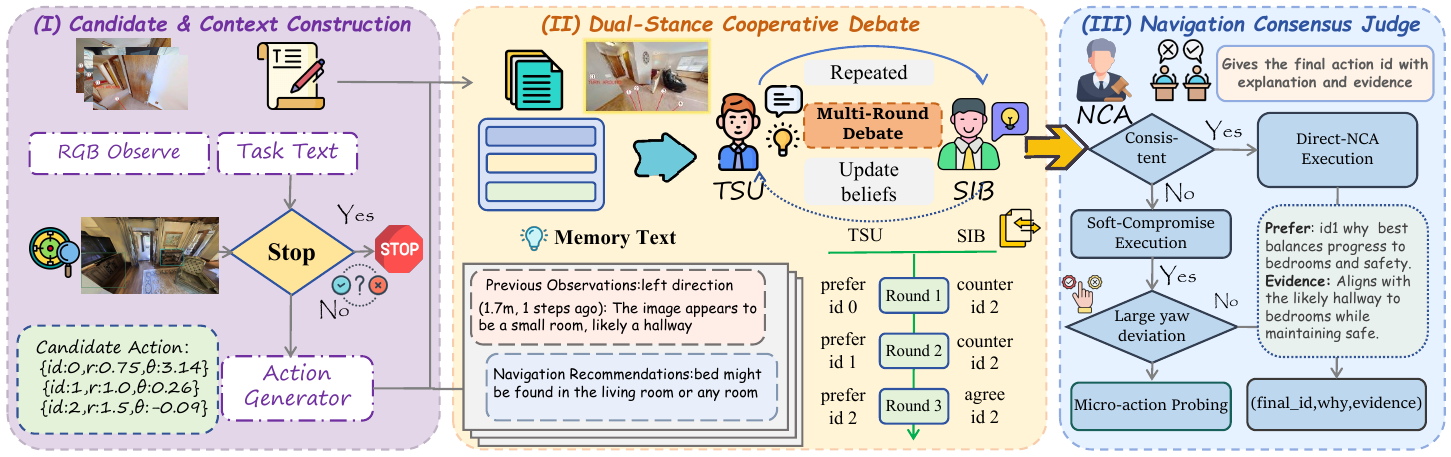}}
    \caption{
        \textbf{DSCD-Nav overview.}
        \textit{(I)} Candidate and context construction packages the goal, optional memory, and evidence cues, and a geometry-pruned set of executable polar action candidates with language descriptions into debate-ready inputs.
        \textit{(II)} TSU and SIB conduct multi-round cooperative debate over the shared candidates, exchanging preferences and cue-grounded evidence while updating beliefs.
        \textit{(III)} The NCA judge arbitrates the debate to produce a final action with a concise rationale and supporting evidence, after which the system executes it directly under consensus or applies bounded geometric soft-compromise with micro-probing when disagreement persists, and the angular gap is within a preset threshold.
        }
    \label{fig:dscd_overview}
  \end{center}
\end{figure*}

\subsection{Multi-Agent Debate and Consensus Decision}
In multi-agent debate, voting, and judge-based frameworks \citep{NEURIPS2024_32e07a11,choi2025debatevoteyieldsbetter}, agents with different stances exchange reasons and evidence over the same query, and a consensus or arbitration module aggregates their views \citep{wang2025marsefficientmultiagentcollaboration} to reduce overconfidence \citep{lin-hooi-2025-enhancing} and make disagreements explicit \citep{liang-etal-2024-encouraging}. In embodied and robotic settings, work on LLM-driven multi-agent coordination \citep{zhang-etal-2025-towards-efficient} and long-horizon planning \citep{NEURIPS2024_7d6e85e8,wang2025multiagentllmactorcriticframework} informs action decision-making. However, existing debate and consensus methods are mostly confined to textual reasoning or high-level plans, with limited focus on candidate-action selection in 3D navigation, where action-level cross-checking and arbitration remain underexplored.

\section{Method}
\paragraph{Task Definition.}
We consider \textit{zero-shot 3D indoor object navigation},
where an agent receives first-person observations $o_t$ and a target
semantic text $g$ at each discrete step $t$. Let $s_t$ denote the true
environment state. After executing action $a_t$, the state evolves as
$s_{t+1}\sim\mathcal{T}(s_t,a_t)$. An episode terminates when the agent
executes \textsc{Stop} or reaches the maximum step budget, and succeeds
if \textsc{Stop} is called within the success radius of the target object.

\paragraph{Method Overview.}
We propose Dual-Stance Cooperative Debate Navigation (DSCD-Nav), as
shown in Fig.~\ref{fig:dscd_overview}. At each step, DSCD-Nav takes the
upstream candidate set $\mathcal{A}_t$ as a unified action interface and
wraps it into debate-ready candidate cards $\mathcal{C}_t$. Together
with the goal text $g$ and optional evidence context $\mathcal{E}_t$,
they form the per-step input $X_t=(o_t,g,\mathcal{C}_t,\mathcal{E}_t)$.
Given the same candidates, a Task--Scene Understanding (TSU) agent
prioritizes goal progress and scene-layout cues, while a
Safety--Information Balancing (SIB) agent audits traversability,
and observation utility. Through multi-round interaction, TSU and SIB
exchange supporting and rebuttal evidence, producing a structured debate
trace $\mathcal{D}_t$ rather than independent scores. The trace is then
arbitrated by a Navigation Consensus Arbitration (NCA) agent to select
the final action. DSCD-Nav directly executes the arbitrated action under
consensus, or performs conservative soft-compromise micro-probing when
near-direction disagreement remains.

\begin{figure*}[ht]
  \begin{center}
    \centerline{\includegraphics[width=\textwidth]{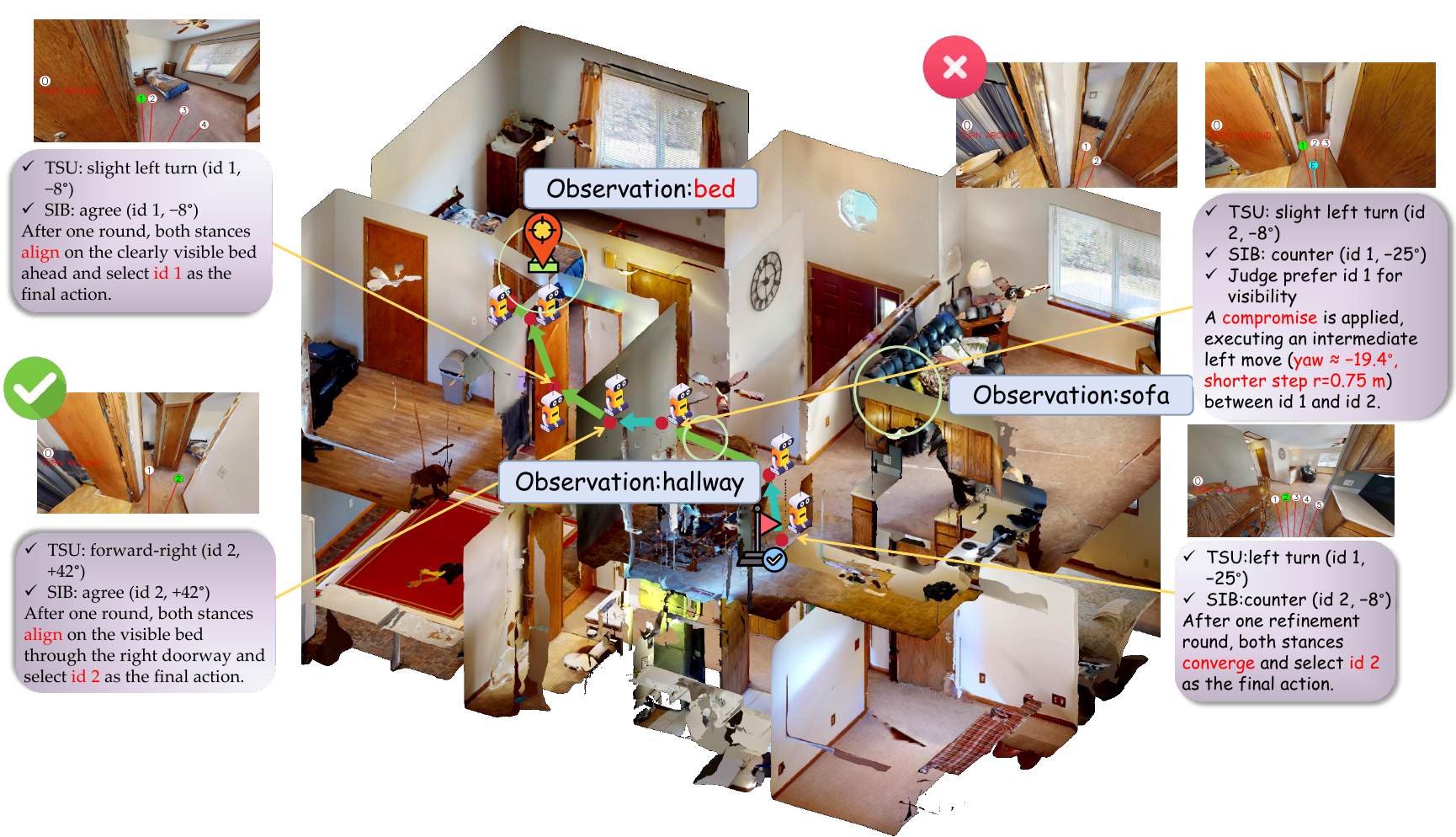}}
    \caption{
        Qualitative case study on HM3Dv2 ObjectNav. We overlay the executed trajectory on the reconstructed 3D scene and annotate several decision points with the corresponding TSU and SIB preferences and the NCA arbitration result. The trace highlights DSCD-Nav’s ability to resolve disagreement at visually ambiguous viewpoints and, when needed, trigger a conservative micro-probing step before committing to the executed action.
    }
    \label{fig:case_hm3dv2}
  \end{center}
\end{figure*}

\subsection{Candidate Action and Context Interface}
To avoid assuming a fixed discrete action set while keeping multi-round debate prompts compact, we follow WMNav~\cite{11246684} and use a geometry-aware candidate generator
that represents executable actions in continuous polar form. At each
step $t$, the generator proposes feasible actions according to local
navigability and collision constraints:
\begin{equation}
    \mathcal{A}_t=\{a_i^t\}_{i=1}^{N_t},\quad a_i^t=(r_i^t,\theta_i^t),
\end{equation}
where $(r_i^t,\theta_i^t)$ are the forward distance and yaw offset relative to the agent's current heading of candidate $i$ at step $t$, and $N_t$ is the number of candidates.

DSCD-Nav introduces a lightweight candidate representation
interface $\Phi(\mathcal{A}_t)\rightarrow \mathcal{C}_t$, which
converts feasible actions into standardized candidate representations
with stable within-step identities:
\begin{equation}
    \mathcal{C}_t=\{c_i^t\}_{i=1}^{N_t},\quad c_i^t=(id_i^t,r_i^t,\theta_i^t,d_i^t),
\end{equation}
where $id_i^t$ is a step-local identity,
$d_i^t$ is a machine-readable direction description, and
$N_t$ is the number of candidates.
The operator $\Phi(\cdot)$ does not modify candidate geometry
or scores; it only standardizes the candidates for
cross-stance comparison.
With optional evidence $\mathcal{E}_t$, the per-step input is:
\begin{equation}
X_t=(o_t,g,\mathcal{C}_t,\mathcal{E}_t).
\end{equation}

\subsection{Dual-Stance Cooperative Debate}
\label{sec:dual_stance_debate}

Prior analyses suggest that homogeneous debate may reinforce
shared misconceptions, while symmetric interaction alone does
not necessarily produce corrective updates
~\cite{NEURIPS2024_32e07a11,choi2025debatevoteyieldsbetter}.
Motivated by these observations, DSCD-Nav replaces one-shot
candidate scoring with an asymmetric cooperative debate over a
shared candidate set.
TSU maintains a goal-directed action hypothesis, while SIB
audits whether it is sufficiently supported by traversability, visibility, and observation evidence.
The two roles are complementary and non-exchangeable.

At round $k$, both agents read the accumulated history
$\mathcal{H}^{(k-1)}$ and revise their candidate preferences
based on previous rationales and evidence.
TSU supports a goal-directed proposal, whereas SIB may agree or
issue a counter.
A valid counter must identify a concrete candidate-level
deficiency and provide an executable alternative, thereby
constraining unsupported revisions.

We describe the exchanged evidence using a lightweight
support--rebuttal relation:
\begin{equation}
\mathcal{R}^{(k)}
=
\mathcal{R}_{\mathrm{sup}}^{(k)}
\cup
\mathcal{R}_{\mathrm{att}}^{(k)},
\end{equation}
where $\mathcal{R}_{\mathrm{sup}}^{(k)}$ supports candidate
actions and $\mathcal{R}_{\mathrm{att}}^{(k)}$ contains grounded
rebuttals to previous preferences or evidence.

The accumulated interaction updates the agents' implicit beliefs
and action preferences across debate rounds. Supporting evidence
encourages convergence, while unresolved rebuttals preserve
structured disagreement for NCA arbitration. This process
promotes correction while limiting harmful revisions; the
corresponding correction--subversion analysis is provided in
Supplementary Material~B.

\paragraph{TSU-Agent: Task--Scene Understanding.}
TSU serves as the goal-progress proposer.
It combines target semantics, the current observation,
scene-layout cues, navigation history, and optional evidence
$\mathcal{E}_t$ to select the candidate most likely to advance
the target search.

At round $k$, TSU outputs:
\begin{equation}
\left(
id_{\mathrm{TSU}}^{(k)},
w_{\mathrm{TSU}}^{(k)},
ev_{\mathrm{TSU}}^{(k)}
\right)
=
f_{\mathrm{TSU}}
\left(
X_t,\mathcal{H}^{(k-1)}
\right),
\end{equation}
where $id_{\mathrm{TSU}}^{(k)}$ is the preferred candidate,
$w_{\mathrm{TSU}}^{(k)}$ is a concise rationale, and
$ev_{\mathrm{TSU}}^{(k)}$ contains supporting visual or
contextual evidence.

\paragraph{SIB-Agent: Safety--Information Balancing.}
SIB serves as a goal-deemphasized safety--information auditor.
Given candidates filtered by upstream geometric feasibility
checks, it further evaluates decision-level risk, traversability
uncertainty, limited visibility, and premature commitment. SIB
returns \texttt{counter} only when it identifies a grounded
deficiency and an executable alternative; otherwise, it returns
\texttt{agree}.

At round $k$, SIB outputs
\begin{equation}
\begin{split}
\big(
dec^{(k)},
id_{\mathrm{SIB}}^{(k)},
w_{\mathrm{SIB}}^{(k)},
ev_{\mathrm{SIB}}^{(k)}
\big)
=
f_{\mathrm{SIB}}
\left(
X_t,
id_{\mathrm{TSU}}^{(k)},
\mathcal{H}^{(k-1)}
\right),
\end{split}
\end{equation}
where
\begin{equation}
dec^{(k)}
\in
\left\{
\texttt{agree},
\texttt{counter}
\right\}.
\end{equation}
When $dec^{(k)}=\texttt{counter}$,
$id_{\mathrm{SIB}}^{(k)}$ denotes the executable alternative,
while $w_{\mathrm{SIB}}^{(k)}$ and
$ev_{\mathrm{SIB}}^{(k)}$ provide its rationale and evidence
grounded in local traversability, risk reduction, or expected
observability.
When $dec^{(k)}=\texttt{agree}$, no alternative candidate is
introduced.

\begin{figure*}[ht]
  \begin{center}
    \centerline{\includegraphics[width=\textwidth]{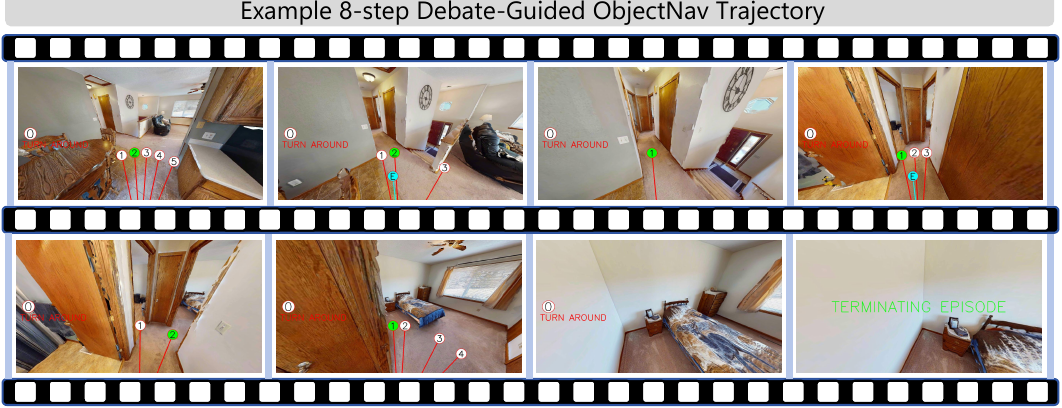}}
    \caption{
        An 8-step, debate-guided HM3Dv1 ObjectNav episode from the agent’s first-person view, illustrating sequential observations, executed actions, and successful termination at the target.
    }
    \label{fig:fp_trace}
  \end{center}
\end{figure*}

\paragraph{Structured Debate Trace.}
Let $K_t\leq K$ denote the actual number of debate rounds
at step $t$. The debate terminates once the two stances reach
agreement; otherwise, it continues until the maximum number
of rounds $K$ is reached. The resulting interaction trace is
summarized as:
\begin{equation}
\label{eq:dataset_t}
\begin{split}
\mathcal{D}_t
=
\Big\{
\big(
&id_{\mathrm{TSU}}^{(k)},
w_{\mathrm{TSU}}^{(k)},
ev_{\mathrm{TSU}}^{(k)},\\
&dec^{(k)},
id_{\mathrm{SIB}}^{(k)},
w_{\mathrm{SIB}}^{(k)},
ev_{\mathrm{SIB}}^{(k)}
\big)
\Big\}_{k=1}^{K_t}.
\end{split}
\end{equation}
The trace preserves candidate-aligned preference revisions and
evidence across rounds and is passed to NCA for final
arbitration.

\subsection{Two-Mode Execution Policy}

To unify dual-stance outputs into an executable action, we
introduce NCA-Agent to arbitrate the debate trace and trigger
a two-mode execution policy.

\paragraph{NCA-Agent: Evidence-Aware Arbitration.}

NCA-Agent reads $\mathcal{C}_t$ and the full debate trace
$\mathcal{D}_t$ to produce a stance-agnostic, evidence-aware
final decision:
\begin{equation}
(id^\star,w^\star,ev^\star)=
f_{\text{NCA}}(g,\mathcal{C}_t,\mathcal{D}_t).
\end{equation}
Here, $id^\star$ is the final high-level action identity, and
$w^\star$ and $ev^\star$ are summaries distilled
from both stances' arguments. We denote the geometry
associated with $id^\star$ as
$a^\star=(r^\star,\theta^\star)$. NCA does not simply pick
a \emph{winner}. Instead, it evaluates proposals under a unified
set of criteria aligned with task progress, safety, and
information considerations, producing an interpretable
consensus action. When disagreement remains at the final
round, NCA arbitrates between the two final stance
preferences and selects one as $id^\star$.

\paragraph{Mode Selection.}

We define a consensus indicator based on agreement at the
actual terminal debate round:
\begin{equation}
\label{eq:consistency_indicator}
\mathbb{I}^{\text{cons}}_{t}
=
\mathbb{I}
\left[
dec^{(K_t)}=\text{agree}
\right],
\end{equation}
where $\mathbb{I}[\cdot]$ is the indicator function
(1 if the condition holds, 0 otherwise).

When disagreement remains, $\theta^\star$ denotes the yaw
angle of the NCA-arbitrated action, while
$\theta^{\text{alt}}$ denotes the yaw angle of the
non-selected final stance preference. Their signed shortest
angular difference is
$\Delta\theta
=
\mathrm{wrap}
\left(
\theta^{\text{alt}}-\theta^\star
\right)
\in(-\pi,\pi].$
The execution mode is selected as:
\begin{equation}
\label{eq:dual_mode}
\begin{cases}
\text{Mode B},
&
\text{if }
\mathbb{I}_t^{\text{cons}}=0
\ \land\
|\Delta\theta|\le\alpha,
\\
\text{Mode A},
&
\text{otherwise},
\end{cases}
\end{equation}
where $\alpha$ is the compromise threshold. Mode A directly
executes the NCA-arbitrated action, while Mode B applies a
conservative soft-compromise under geometrically close
disagreement.

\paragraph{Mode A: Direct NCA Execution.}

Mode A directly executes the NCA action when the final
preferences agree or when their directional disagreement is
too large for a local probe:
\begin{equation}
(r_{\text{exec}},\theta_{\text{exec}})
=
(r^\star,\theta^\star).
\end{equation}

\paragraph{Mode B: Soft-Compromise and Micro-Probing.}

When disagreement persists after multi-round debate, we
check whether the two final preferences are geometrically
close enough for a conservative compromise. We define:
\begin{equation}
a^\star=(r^\star,\theta^\star),\quad
a^{\text{alt}}=(r^{\text{alt}},\theta^{\text{alt}}),
\end{equation}
where $a^{\text{alt}}$ denotes the non-selected final stance
preference after NCA arbitration. If
$|\Delta\theta|\le\alpha$, we apply a geometric
soft-compromise: the
high-level identity remains $id^\star$, while the low-level
execution is adjusted via shortened forward motion and a
slight yaw bias toward the alternative direction:
\begin{equation}
r_{\text{exec}}
=
\beta_r\,r^\star,
\qquad
\theta_{\text{exec}}
=
\theta^\star+\beta_\theta\Delta\theta.
\end{equation}
We implement Mode B as a micro-probing step parameterized
by $(\beta_r,\beta_\theta)$, where $\beta_r$ scales the
forward motion and $\beta_\theta$ interpolates the yaw toward
the alternative direction. This micro-probing step keeps the
execution close to $a^\star$ while inducing a controlled
viewpoint change to acquire disambiguating evidence before
the next debate--arbitration step. We use
$(\beta_r,\beta_\theta)=(\tfrac{1}{2},\tfrac{1}{3})$ by
default and report a sensitivity analysis in Supplementary Material C.1. If the directional gap exceeds the
threshold $\alpha$, we directly execute the NCA-arbitrated
action $a^\star$ using Mode A.

\begin{table*}[ht]
  \centering
  \caption{Comprehensive comparison with state-of-the-art methods on ObjectNav benchmarks. TF denotes training-free, and ZS denotes zero-shot. Our method uses the lightweight \texttt{Gemini-2.5-Flash-Lite} as the base VLM.}
  \label{tab:sota_objectnav}
  \begin{small}
  \begin{tabular}{lcc cc cc cc}
    \toprule
    Method & ZS & TF &
    \multicolumn{2}{c}{HM3Dv1} &
    \multicolumn{2}{c}{HM3Dv2} &
    \multicolumn{2}{c}{MP3D} \\
    \cmidrule(lr){4-5}\cmidrule(lr){6-7}\cmidrule(lr){8-9}
    & & &
    SR(\%) $\uparrow$ & SPL(\%) $\uparrow$
    & SR(\%) $\uparrow$ & SPL(\%) $\uparrow$
    & SR(\%) $\uparrow$ & SPL(\%) $\uparrow$ \\
    \midrule

    \rowcolor{gray!15}
    SemEXP~\cite{NEURIPS2020_2c75cf26}
    & $\times$ & $\times$
    & -- & --
    & -- & --
    & 36.0 & 14.4 \\

    ZSON~\cite{NEURIPS2022_d0b8f0c8}
    & $\times$ & $\times$
    & 25.5 & 12.6
    & -- & --
    & 15.3 & 4.8 \\

    \rowcolor{gray!15}
    PONI~\cite{Ramakrishnan_2022_CVPR}
    & $\times$ & $\times$
    & -- & --
    & -- & --
    & 31.8 & 12.1 \\

    PSL~\cite{10.1007/978-3-031-73254-6_10}
    & $\times$ & $\times$
    & 42.4 & 19.2
    & -- & --
    & -- & -- \\

    \rowcolor{gray!15}
    Habitat-Web~\cite{Ramrakhya_2022_CVPR}
    & $\checkmark$ & $\times$
    & 41.5 & 16.0
    & -- & --
    & 31.6 & 8.5 \\

    ProcTHOR-ZS~\cite{NEURIPS2022_27c546ab}
    & $\checkmark$ & $\times$
    & 13.2 & 7.7
    & -- & --
    & -- & -- \\

    \rowcolor{gray!15}
    Pixel-Nav~\cite{10610499}
    & $\checkmark$ & $\times$
    & 37.9 & 20.5
    & -- & --
    & -- & -- \\

    SGM~\cite{Zhang_2024_CVPR}
    & $\checkmark$ & $\times$
    & 60.2 & 30.8
    & -- & --
    & 37.7 & 14.7 \\

    \rowcolor{gray!15}
    VLFM~\cite{10610712}
    & $\checkmark$ & $\times$
    & 52.5 & 30.4
    & 63.6 & 32.5
    & 36.4 & 17.5 \\

    ESC~\cite{pmlr-v202-zhou23r}
    & $\checkmark$ & $\checkmark$
    & 39.2 & 22.3
    & -- & --
    & 28.7 & 14.2 \\

    \rowcolor{gray!15}
    OpenFMNav~\cite{kuang-etal-2024-openfmnav}
    & $\checkmark$ & $\checkmark$
    & 51.9 & 24.4
    & -- & --
    & 37.2 & 15.7 \\

    InstructNav~\cite{long2024instructnav}
    & $\checkmark$ & $\checkmark$
    & -- & --
    & 58.0 & 20.9
    & -- & -- \\

    \rowcolor{gray!15}
    SG-Nav~\cite{NEURIPS2024_098491b3}
    & $\checkmark$ & $\checkmark$
    & 54.0 & 24.9
    & 49.6 & 25.5
    & 40.2 & 16.0 \\

    OneMap~\cite{11128393}
    & $\checkmark$ & $\checkmark$
    & 55.8 & 37.4
    & -- & --
    & -- & -- \\

    \rowcolor{gray!15}
    TopV-Nav~\cite{zhong2025topvnavunlockingtopviewspatial}
    & $\checkmark$ & $\checkmark$
    & 52.0 & 28.6
    & -- & --
    & 35.2 & 16.4 \\

    CogNav~\cite{Cao_2025_ICCV}
    & $\checkmark$ & $\checkmark$
    & 72.5 & 26.2
    & -- & --
    & 46.6 & 16.1 \\

    \rowcolor{gray!15}
    CoS~\cite{cos2026}
    & $\checkmark$ & $\checkmark$
    & 55.9 & 29.1
    & --  & --
    & 37.6 & 17.6 \\

    PanoNav~\cite{panonav2026}
    & $\checkmark$ & $\checkmark$
    & 43.5 & 23.7
    & -- & --
    & -- & -- \\

    \midrule

    \rowcolor{gray!15}
    \textbf{DSCD (Ours)}
    & $\checkmark$ & $\checkmark$
    & \textbf{75.6} & \textbf{40.1}
    & \textbf{73.0} & \textbf{38.7}
    & \textbf{47.8} & \textbf{24.2} \\

    \bottomrule
  \end{tabular}
  \end{small}
\end{table*}

\section{Experiment}

\subsection{Experiment Setup}
\paragraph{Datasets.}
We evaluate in Habitat on the validation splits of HM3Dv1~~\cite{NEURIPS_DATASETS_AND_BENCHMARKS2021_34173cb3}, HM3Dv2~~\cite{Yadav_2023_CVPR}, and MP3D ObjectNav~~\cite{8374622}.

\paragraph{Implementation Details.}
The action set includes \texttt{stop}, \texttt{rotate}, and \texttt{move\_forward} with forward distance sampled from a continuous range $[0.5\,\text{m},\,1.7\,\text{m}]$  to balance long strides and local probing. DSCD‑Nav uses  the lightweight \texttt{Gemini‑2.5‑Flash‑Lite} as its vision–language backbone. For comparisons with discrete-action baselines, we provide the step-conversion protocol and discretized-action evaluation in
Supplementary Material A.3.

\paragraph{Evaluation Metrics.}
We evaluate using Success Rate (SR), Success weighted by Path Length (SPL), and the Area Overlap Redundancy Index (AORI). SR is the fraction of episodes where the agent stops successfully near the target. SPL measures path efficiency by comparing shortest-path length to the executed path length on successful episodes. AORI quantifies exploration redundancy by penalizing revisits to previously covered areas (lower is better). Formal definitions are provided in Supplementary Material A.5.

\subsection{Methods Comparison}
\paragraph{Qualitative Analysis.}
As shown in Fig.~\ref{fig:case_hm3dv2}, DSCD-Nav avoids premature commitment at visually ambiguous junctions by balancing TSU and SIB stances through debate and NCA arbitration, with conservative micro-probing when needed, reducing detours along the trajectory. Fig.~\ref{fig:fp_trace} further provides a first-person trace, illustrating how debate-informed decisions evolve over sequential observations and lead to timely termination once target evidence is confirmed.
\paragraph{Quantitative Analysis.}
We evaluate on ObjectNav in HM3Dv1, HM3Dv2, and MP3D, and
compare with other methods in Table~\ref{tab:sota_objectnav}.
DSCD-Nav achieves higher SR and SPL across all benchmarks,
reaching 73.0\% SR and 38.7\% SPL on HM3Dv2. These results
demonstrate that structured dual-stance decision making enables
more robust candidate selection and generalizes across datasets.

Furthermore, we compare with representative training-free
VLM-based baselines on HM3Dv2 (ObjectNav, val;
Table~\ref{tab:tf_benchmarks}(a)) and HM3Dv1 (GOAT, val unseen;
Table~\ref{tab:tf_benchmarks}(b)).
DSCD-Nav improves over previous VLM-based methods, achieving
73.0\% SR and 38.7\% SPL on ObjectNav and 40.1\% SR and
32.3\% SPL on GOAT. It also reduces AORI to 39.2\% and 41.9\%,
respectively, indicating more stable progress and reduced
exploration redundancy in long-horizon navigation.
\begin{table}[ht]
  \centering
  \caption{Comparison of training-free VLM-based navigation methods on ObjectNav and GOAT benchmarks.}
  \label{tab:tf_benchmarks}
  \begin{small}
  \textbf{(a) ObjectNav benchmark}\\
  \begin{tabular}{lccc}
    \toprule
    Method & SR (\%)$\uparrow$ & SPL (\%)$\uparrow$ & AORI (\%)$\downarrow$ \\
    \midrule
    Prompt-only        & 29.8 & 10.7 & --  \\
    PIVOT              & 24.6 & 10.6  & 63.3 \\
    VLMNav             & 51.6 & 18.3  & 61.5 \\
    DORAEMON           & 62.0 & 23.0 & 50.1 \\
    \midrule
    \textbf{DSCD (Ours)}      & \textbf{73.0} & \textbf{38.7} & \textbf{39.2} \\
    \bottomrule
  \end{tabular}

  \textbf{(b) GOAT benchmark}\\
  \begin{tabular}{lccc}
    \toprule
    Method & SR (\%)$\uparrow$ & SPL (\%)$\uparrow$ & AORI (\%)$\downarrow$ \\
    \midrule
    Prompt-only        & 11.3 & 3.7 & --  \\
    PIVOT & 8.3  & 3.8 & 64.9 \\
    VLMNav & 22.1 & 9.3 & 63.6 \\
    DORAEMON           & 24.3 & 10.3 & 56.9 \\
    \midrule
    \textbf{DSCD (Ours)}      & \textbf{40.1} & \textbf{32.3} & \textbf{41.9} \\
    \bottomrule
  \end{tabular}

  \end{small}
\end{table}

\subsection{Real-World Robot Experiments}

\begin{figure}[ht]
    \centering
    \includegraphics[width=\columnwidth]{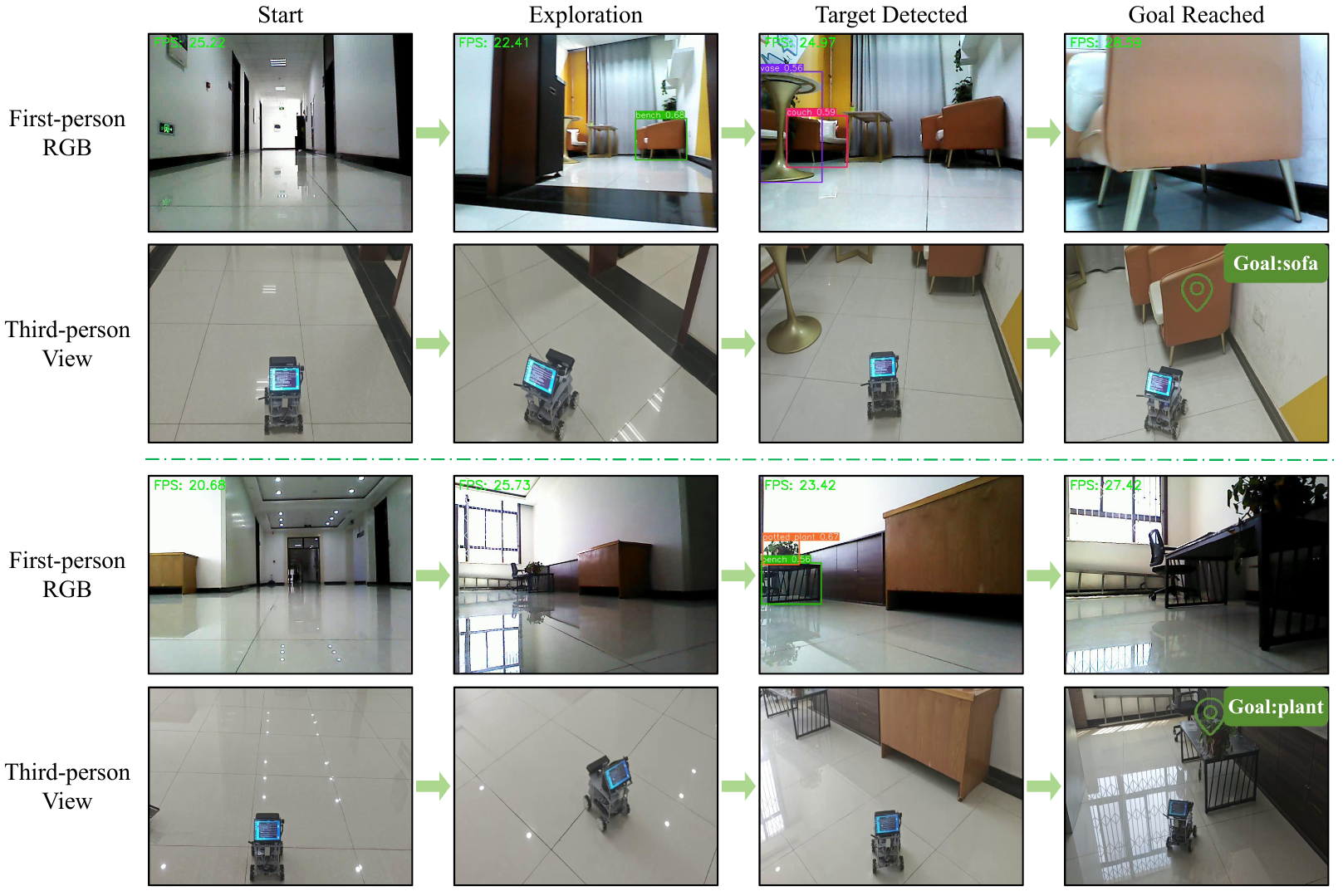}
    \caption{
    Real-world ObjectNav testing. Sequences show first-person RGB observations and third-person views for sofa and plant targets from start to goal.
    }
    \label{fig:real_world}
\end{figure}

To examine real-world deployability, we deploy DSCD-Nav on a Wheeltec R550 mobile robot with an Orbbec Astra Pro Plus RGB-D camera and an NVIDIA Jetson Orin NX Super 16GB onboard computing platform. The system uses a local computer for communication and control, and a remote NVIDIA A6000 server for decision inference.
As shown in Fig.~\ref{fig:real_world}, the robot successfully reaches \textit{sofa} and \textit{plant} targets in real indoor scenes, showing that DSCD-Nav can transfer to physical navigation under real sensor noise, viewpoint changes, and imperfect detections.

\subsection{Ablation Study}
We conduct comprehensive ablation studies to validate the design choices of DSCD-Nav, with results summarized in Table \ref{ablation_hm3dv2}. 
\begin{table}[ht]
  \centering
  \caption{Ablation studies of DSCD modules, debate rounds, and backbone VLMs on HM3Dv2 ObjectNav benchmark.}
  \label{ablation_hm3dv2}
  \begin{small}
  \begin{tabular}{lccc}
    \toprule
    Method / Configuration & SR(\%)$\uparrow$ & SPL(\%)$\uparrow$ & AORI(\%)$\downarrow$ \\
    \midrule

    \rowcolor{gray!15}
    \multicolumn{4}{c}{\textbf{(a) Ablation of DSCD modules}} \\
    \midrule
        TSU-only           & 62.4   & 27.6   & 49.5   \\
        SIB-only           & 56.5   & 25.9   & 44.1   \\
        w/o NCA-Agent      & 67.8   & 33.8   & 43.2   \\
        w/o micro-probing  & 71.2   & 36.9   & 41.0   \\
        DSCD (Ours, default) & 73.0 & 38.7   & 39.2   \\
    \midrule

    \rowcolor{gray!15}
    \multicolumn{4}{c}{\textbf{(b) Ablation of debate rounds}} \\
    \midrule
    rounds = 1             & 51.3   & 30.8 & 45.0   \\
    rounds = 2             & 65.6   & 36.7 & 41.0   \\
    rounds = 3 (default)   & 73.0   & 38.7 & 39.2   \\
    rounds = 4             & 73.5   & 38.8 & 38.9   \\
    \midrule

    \rowcolor{gray!15}
    \multicolumn{4}{c}{\textbf{(c) Ablation of different VLMs}} \\
    \midrule
    Gemini-2.5-Flash       & 80.1 & 41.1 & 36.1 \\
    Gemini-2.5-Flash-Lite  & 73.0 & 38.7 & 39.2 \\
    Qwen3-VL-8B-Instruct  & 67.8 & 35.1 & 44.7 \\
    GPT-4.1-nano           & 58.7 & 29.9 & 50.5   \\
    \bottomrule
  \end{tabular}
  \end{small}
\end{table}

\paragraph{(a) Core components.}
Removing either TSU or SIB consistently degrades performance.
Compared with the full model (73.0\% SR), TSU-only and SIB-only
reduce SR to 62.4\% and 56.5\%, respectively, demonstrating their
complementary roles in decision making. Removing NCA further
reduces SR to 67.8\%, showing the importance of unified
arbitration under disagreement. Disabling micro-probing reduces
low-cost verification during disagreement and slightly decreases
SPL while increasing AORI.

\paragraph{(b) Debate rounds.} Table~\ref{ablation_hm3dv2}(b) shows that using more debate rounds generally improves decision quality by enabling richer interaction and evidence refinement. However, the marginal gains saturate as rounds increase, while inference cost grows, so we adopt a default setting that balances performance and efficiency.

\paragraph{(c) Choice of VLM.}
As shown in Table~\ref{ablation_hm3dv2}(c), performance varies
across VLM backbones, but DSCD-Nav remains effective with the
lightweight \texttt{Gemini-2.5-Flash-Lite}, achieving 73.0\% SR
and 38.7\% SPL. This indicates that the gains come from the
decision structure rather than relying on a specific backbone.
\begin{table}[ht]
  \centering
  \caption{Comparison of disagreement-related debate metrics and $\Delta$SPL across Same-Stance Cooperative Debate (SSCD) and DSCD configurations. Metric calculation formulas are provided in Supplementary Material A.5}
  \label{consensus_stats}
  \begin{small}
  \setlength{\tabcolsep}{5pt}
  \begin{tabular}{lcccc}
    \toprule
    Config &
    DR(\%) &
    JOR(\%) &
    MPTR(\%) &
    $\Delta$SPL$\uparrow$ \\
    \midrule
    SSCD, rounds=3     & 5.0  & 60.0 & 20.0 & --  \\
    DSCD, rounds=1     & 41.0 & 43.0 & 41.0 & 12.5  \\
    DSCD, rounds=2     & 21.4 & 45.5 & 63.6 & 18.4 \\
    DSCD, rounds=3     & 15.6 & 50.0 & 20.0 & 20.3  \\
    \bottomrule
  \end{tabular}
  \end{small}
\end{table}

To interpret DSCD-Nav at the process level, Table~\ref{consensus_stats} summarizes key disagreement statistics, including the Disagreement Rate (DR) over candidate preferences, the Judge Override Rate (JOR) during arbitration, and the Micro-Probing Trigger Rate (MPTR) when final-round stances disagree. Disagreements concentrate at high-uncertainty steps and reflect explicit candidate conflicts rather than random noise. Arbitration and micro-probing then convert these conflicts into more robust actions, consistent with the gains in SPL.

\section{Conclusion}
In this paper, we propose DSCD-Nav, a training-free dual-stance cooperative debate framework where TSU and SIB cross-check shared candidate actions under partial observability. NCA arbitration integrates both stances into a final action, with lightweight micro probing triggered under persistent near-direction disagreement to validate competing preferences at low risk. Experiments on ObjectNav benchmarks demonstrate consistent improvements in success and path efficiency with reduced exploration redundancy, while ablations verify the contribution of each component. Real-world robot experiments further validate the effectiveness of DSCD-Nav in physical environments.




\bibliography{aaai2027}


\end{document}